\documentclass{article}
\usepackage{ijcai17}
\usepackage{hyperref}
\usepackage{amsmath}
\usepackage{times}
\usepackage{graphicx}

\usepackage{array}

\usepackage{url}
\usepackage{paralist}

\newcolumntype{M}[1]{>{\arraybackslash}m{#1}}
\newcolumntype{N}{@{}m{0pt}@{}}

\title{Interactions as Social Practices: towards a formalization}
\author{F.P.M. Dignum\\Utrecht University\\f.p.m.dignum@uu.nl}

\begin{document}

\maketitle

\begin{abstract}
Multi-agent models are a suitable starting point to model complex social interactions. However, as the complexity of the systems increase, we argue that novel modeling approaches are needed that can deal with inter-dependencies at different levels of society, where many heterogeneous parties (software agents, robots, humans) are interacting and reacting to each other. 
In this paper, we present a formalization of a social framework for agents based in the concept of \emph{Social Practices} as high level specifications of normal (expected) behavior in a given social context. We argue that social practices facilitate the practical reasoning of agents in standard social interactions.
%
%
%
\end{abstract}

\section{Introduction and Motivation}

Interactions do not exist in a vacuum, but are surrounded by many social and physical constructs that shape and constraint that interaction \cite{Argyle}. Understanding and modeling the context of interactions is essential in the design of systems that are both realistic and computationally feasible. In fact, if context is not properly considered  \emph{``the patterns and the outcomes of the interactions are inherently unpredictable, and predicting the behavior of the overall system based on its constituent components is extremely difficult (sometimes impossible) because of the high likelihood of emergent (and unwanted) behavior"}~\cite{jennings:00}. In human societies, we use social practices as means to cope with uncertainty of outcome of interaction. Interactions are embedded in a broad network of societal and institutional contexts, and social practices provide means to deal with this complexity. Endowing agents with means to represent and reason with social practices, will enable smooth, flexible, context-aware human-agent interaction.
%
%

Applied to human-agent interactions, such as companion robots, human-agent-robot teamwork, or persuasive applications, social practices can simplify the deliberation of the agent in complex contexts. In these domains, the agent is expected to support the user but there are no predefined protocols or fixed objectives, so agent actions must be based on a close observation of the situation and evaluation of possible user aims, and reaction to user actions and environment changes. This is exactly the reason for social practices in human interactions. Social practices refer to everyday practices and the way these are typically  performed by (most of) the members in the society. Even though the subject of emergence and evolution of social practices is an important one, for simplicity in this paper we assume social practices to be given and fixed. I.e. we will not elaborate on how the practice come to be, or how they are maintained and shared. Our conceptual agent architecture therefore takes a set of social practices as given, and the focus is on how the agent uses practices in its deliberation and planning.

In this paper, we present a formalization of the theory of social practices as proposed in \cite{shove2012,Reckwitz} to be used as basis for a cognitive architecture. Our 
approach can be seen as a middle way between  using fixed interaction scripts and full free deliberation reactive to user and environment sensing. We build on the concept of landmark \cite{Dignum2007} in the sense, that social practices enable the specification of possible (abstract) plans without needing to represent every single component of the interaction. An agent will then adopt a social practice, based on its evaluation of the context, and \textit{`fill-in the gaps'}, using its own capabilities and according to its own intentions. We provide a logic semantics for social practices, based on modalities to represent expectation, capability, roles, norms and contexts.




\section{Background}
Social practices are accepted ways of doing things, contextual and materially mediated, that are shared between actors and routinized over time \cite{Reckwitz}. They can be seen as patterns which can be filled in by a multitude of single and often unique actions. Through (joint) performance, the patterns provided by the practice are filled out and reproduced.
Each time it is used, elements of the practice, including know-how, meanings and purposes,
are reconfigured and adapted \cite{shove2012}. Therefore the use of social practices includes a constant learning of the individuals using the social practice in ever changing contexts. In this way social practices guide the learning process of agents in a natural way. In \cite{shove2012} the social aspect of social practices is emphasized by giving the social practice center stage in interactions and letting individuals be supporters of the social practice. It shows that social practices are shared (social) concepts. The mere fact that they are shared and jointly created and maintained means that individuals playing a role in a social practice will expect certain behavior and reactions of the other participants in the social practice. Thus it is this aspect that makes the social practices so suitable for use in individual planning in social situations. Because, in this paper, we concentrate on the individual planning, we will not see much of the particular social aspects of the social
practices.

The formal model proposed in this paper is based on modal logics, extending work on agent organizations and landmarks, in particular the Logic of Agent Organizations (LAO) \cite{Dignumigpl} that provides a formal, provable representation of organizations based on the
notions of capability, `bring it about' -- or stit -- \cite{Porn1974}, attempt and responsibility. Landmarks
represent points in the plans of an agent that must always be achieve. This concept has been extensively used in the planning community \cite{}. We use the ADICO grammar proposed by Ostrom to specify rules, norms, and strategies \cite{crawford1995grammar}. ADICO statements
are formed using the following five components: Attribute (or Acting entity), Deontic, aIm (or Intention), Condition , and Or else (or sanction). In ADICO Strategies are statements including only the acting entity, intention, and condition (AIC); Norms
include the acting entity, deontic, intention, and condition (ADIC); and Rules include all 5 components (ADICO). The use of landmarks in the specification of agent plans
and interactions has been proposed by \cite{kumar02,Dignum2007}. Formally, landmarks are conjunctions of logical expressions that are true in a state, representing families of protocols. The level of landmark detail determines the degree of actors freedom.

Related research on work practices and cognitive architectures are the closest to our proposal. Work practice research recognizes the inherent difference between the work flows as described and prescribed by the organization and employee behavior. The work practice model Brahams enables to define the behavior of entities by means of activities and workframes, amongst others~\cite{sierhuis09} but lacks learning capabilities to adjust priorities, and means to distinguish between context and action preconditions. Cognitive architectures \cite{sun2009motivational} use drives as basis to dynamically derive goals during agent interactions. As such, it can capture the motivational complexity of the human mind~\cite{newell1994unified}, but it takes an agent perspective rather than a societal one, such as the one we propose.



\section{Characteristics of Social Practices}\label{sec:spelements}
In Social Sciences, \emph{social practices} (SP) are defined on the basis of materials, meanings and competences~\cite{holtz2014}. In the following we adapt an initial conceptualization of these concepts, as introduced in  \cite{Dignum2015}:\\
\textit{\textbf{Context}}~\cite{Zimmermann:2007}
%
%
    \begin{itemize}
    \item \textit{Roles} describe the competencies and expectations about a certain type of actors~\cite{sunstein96}. Thus a lecturer is expected to deliver the presentation.
    \item \textit{Actors} are all people and autonomous systems involved, that have capability to reason and (inter)act. This indicates the other agents that are expected to fulfill a part in the practice.
    \item \textit{Resources} are objects that are used by the actions in the practice such as seats, projector, screen, etc. So, they are assumed to be available both for standard actions and for the planning within the practice.
    \item \textit{Affordances} are the properties of the context that permit social actions and depend on the match between context conditions and actor characteristics~\cite{gaver1996situating}.
    \item \textit{Places} indicates where all objects and actors are usually located relatively to each other, in space or time: Seats in a lecture theater all face the front of the room, etc.
\end{itemize}
\textit{\textbf{Meaning}}
    \begin{itemize}
    \item \textit{Purpose} determines the social interpretation of actions and of certain physical situations.
    \item \textit{Promotes} indicates the values that are promoted (or demoted, by promoting the opposite) by the social practice. 
    \item \textit{Counts-as} are rules of the type "X counts as Y in C" linking brute facts (X) and institutional facts (Y) in the context (C) \cite{searle1995construction}. E.g., in a voting place, filling out a ballot counts as a vote.
    \end{itemize}
\textit{\textbf{Expectations}}
    \begin{itemize}
    \item \textit{Plan patterns} describe usual patterns of actions~\cite{bresciani2004tropos} defined by the landmarks that are expected to occur.
    \item \textit{Norms} describe the rules of (expected) behavior within the practice, as statements of the form ADIC or ADICO.
    \item \textit{Strategies}  indicate the possible activities that are expected within the practice. Not all activities need to be performed! They are meant as potential courses of action. Strategies are specified as AIC statements 
    \item \textit{Start condition}, or trigger, indicating how SP starts
    \item \textit{Duration}, or End condition, indicating how SP ends
    \end{itemize}
\textbf{\textit{Activities}}
    \begin{itemize}
    \item \textit{Possible actions} describes the expected actions by actors in the social practice
    \item \textit{Requirements} indicate the type of capabilities or competences that the agent is expected to have in order to perform the activities within this practice.
    \end{itemize}


%


Social Practices show some resemblance to agent organizations (see e.g. \cite{OperA}), in the sense that both provide structure to the interactions between agents. However, organizations provide an imposed (top-down) structure, while the social practices form a structure that arises from the bottom up. Thus instead of compliance, interaction patterns in a social practice indicate expectations on the behavior of its actors. Therefore no guarantee is given that behavior will occur (exactly) as expected. This has a large influence on the way the social practices are formally specified, focusing on possibilities and priorities rather than prescribing behavior. Cf. Table \ref{table:sp}.A for an informal description of how to use social practices.
Expectations are given in the form of roles that determine possible and expected behavior and also indicate who is supposed to take initiative at certain points. E.g. the lecturer is supposed to take the initiative to start the lecture. Norms determine the normal patterns of behavior and thus also determine a certain type of expectation, namely the patterns of actions that are allowed or prohibited. The plan patterns also determine partly the sequences of actions that are expected. The different scenes are temporally ordered and have a specific starting and end situation. Thus the whole practice should be fulfilled by following a trace that fits through that plan pattern.
%
%

In the next section we will start defining a formal representation for social practices based on the above observations.

\section{Formal model}\label{sec:logic}
 Because the description of actions and action sequences is important as well as epistemic/doxastic states of the agents we will use a combination of dynamic logic with epistemic logic as the basis for our formalization.\footnote{Given that we only need to represent epistemic and dynamic operators, and not their dynamics, Dynamic epistemic logic \cite{Ditmarsch} is not needed here.}

\subsection{Activities}
We start by formalizing social practice activities in terms of actions and capabilities.
 
Let $Act=\{\alpha_1,...\alpha_n\}$ be the set of basic actions and $Ag=\{a_1,...a_m\}$ be the set of actors. We define a function $Capability: Agx2^{Act}$ that indicates for each actor the set of (basic) actions it is capable to perform. We have $Cap(a,\alpha) {\bf iff} \alpha\in Capability(a)$.\\
We use $[\alpha(a)]\phi$ to denote that if actor $a$ performs action $\alpha$ the result $\phi$ will become true afterwards. We assume the axiom: $([\alpha(a)]{\bf true}) \rightarrow Cap(a,\alpha)$ such that we do not have to specify the capability as a precondition every time we use the action description with an actor.\\
%
%
We can generalise these definitions for unions of actions, parallel actions, sequences of actions and repetitions:\\
The set ${\sl CA}$ of complex action expressions is given as the smallest set closed under:
\begin{itemize}
\item[(i).] $Act \subseteq {\sl CA}$
\item[(ii).] $ \alpha_1 , \alpha_2 \in {\sl CA} \Longrightarrow  \alpha_1
+ \alpha_2 \in {\sl CA}$
\item[(iii).] $ \alpha_1 , \alpha_2 \in {\sl CA} \Longrightarrow  \alpha_1
\& \alpha_2  \in {\sl CA}$
\item[(iv).] $ \alpha_1 , \alpha_2 \in {\sl CA} \Longrightarrow  \alpha_1 ; \alpha_2  \in {\sl CA}$
\item[(v).] $ \alpha \in {\sl CA} \Longrightarrow  \alpha * \in {\sl CA}$
\end{itemize}
Let $\gamma \in \sl CA$ and $A \subseteq Ag$ then $[\gamma(A)]\phi$ denotes that $\phi$ will be true after the group of actors $A$ has executed the complex action $\gamma$. Here we use $\gamma(A)$ as an abbreviation for the following:\\
Let $\gamma\equiv \gamma_1 \circ \gamma_2$ with $"\circ"$ is $"+", "\&"$ or $";"$ or $\gamma\equiv \gamma_1 ; \gamma_2 \wedge \gamma_2\equiv \gamma_1 *$ then
\[\gamma(A) \equiv \exists A',A'': A'\cup A''=A \wedge \gamma_1(A')\circ\gamma_2(A'')\]
%
%
Thus if we indicate that a group $A$ performs an action we explicitly do not differentiate who performs what part of that action. If we want to be more specific we can indicate the specific group performing a sub-action with that sub-action directly. Thus:
\[ (\gamma_1 \circ \gamma_2)(A) \not\equiv \gamma_1(A)\circ\gamma_2(A)\]
We use $act(\gamma)$ to denote the set of basic actions that are part of the complex action $\gamma$.
º
It will be handy to refer to abstract actions such as "an action that achieves $\phi$" or "an action performed by $A$ that achieves $\phi$". For this we introduce the following definitions:\\
Let $\alpha_i \in \sl CA$
\[ \alpha\phi \equiv \cup\alpha_i : [\alpha_i]\phi \]
\[ \alpha(A)\phi \equiv \cup\alpha_i(A) : [\alpha_i(A)]\phi \]
Note that if $\phi=true$ the abstract action refers to any action that can be performed.

We use $DO(a,\alpha)$ (resp. $DONE(a,\alpha)$) to denote that actor $a$ performs action $\alpha$ next (resp. actor $a$ performed action $\alpha$ as the last action). These action modalities can be represented in the semantics of dynamic logic by introducing a selection function on the links that indicate all possible actions in the current state. (see e.g. \cite{Dignumigpl} for details). Again we assume $DO(a,\alpha) \rightarrow Cap(a,\alpha)$ and $DONE(a,\alpha) \rightarrow Cap(a,\alpha)$.\\
We also extend this notation to groups of actors and use $DO(A,\alpha)$ and $DONE(A,\alpha)$ to denote a group $A$ doing an action together. Finally we use $DO(a,\alpha(A))$ and $DONE(a,\alpha(A))$ to denote that $a$ performs actions as part of the group $A$ in order to perform $\alpha$ together.

%
%

Besides actions we will also refer to contexts within which an action can be performed. This will be especially important when effects of an action depend on the context. E.g. the effect of raising your hand can be different in an auction and in a classroom. Intuitively a context stands for a temporal and/or spatially defined  interval within which an action takes place, such as "the lunch break" or "the board room of university". 
%
%
There is a vast amount of literature on reasoning in context. We will not tap on this literature, because we mainly use contexts (in this paper) as reference structures needed to specialize interaction expectations and beliefs. Every social practice can also function as a context by using only those parts that give temporal and spatial restrictions on the actions and agents.\\
We assume a set $C=\{c_1,...,c_n\}$ of special context constants (or names). Let $SP$ be the set of social practice identifiers then: $SP \subset C$. (Thus every social practice is a context, but not every context is a social practice). We explicitly do not have closeness of this set under combinations of contexts in any way! We also do not assume that any combination of specifications of temporal and/or spatial intervals defines a context. However, we do have that with every element $c_i \in C$ there is associated a formula $\Psi_i$ written as $con(c_i,\Psi_i)$ such that:
\[ active(a,\gamma, c_i) \equiv \Psi_i(a,\gamma)\]
meaning that context $c_i$ is active for actor $a$ while performing action $\gamma$ if the conditions associated with that context ($\Psi$) hold for $a$ performing $\gamma$.\footnote{This is a very simplified version of contexts such as is developed in \cite{GrossiDignum}}.If $c_i$ is a social practice then:
\[\Psi_i \rightarrow Sc_{c_i} \wedge \neg D_{c_i}\]
where $Sc_{c_i}$ is the start condition of $c_i$ and $D_{c_i}$ is the duration or end condition.\\
In general it can be the case that more than one context is active when an actor performs an action. This is not always a problem, but especially for concepts like expectations we want to have a unique context to narrow down (disambiguate) which expectations are relevant. Therefore we also assume there is a function $salient: CxC \rightarrow C$ that indicates for any pair of contexts which is the most salient. The following two restrictions hold for this function:
\[active(a,\gamma,c_1) \wedge \neg active(a,\gamma,c_2) \rightarrow salient(c_1,c_2)=c_1\]
\begin{gather*}
active(a,\gamma,c_1) \wedge active(a,\gamma,c_2) \wedge con(c_1,\Psi_1) \wedge con(c_2,\Psi_2) \\
\wedge (\Psi_1(a,\gamma)\rightarrow \Psi_2(a,\gamma))\rightarrow salient(c_1,c_2)=c_1
\end{gather*}
the first restriction states that only active contexts can be salient. The second states that more specific contexts are salient.\\
Finally we define $Salient(a,\gamma,c_i)$ as:
\[active(a,\gamma,c_i) \wedge \neg\exists c_j: salient(c_i,c_j)=c_j\]
which gives the most salient context that is active for actor $a$ when performing $\gamma$.

\subsection{Beliefs and Assumptions}
From epistemic logic we will use especially the operators for "everyone in the set of actors $A$ believes" ($EB_A$) and "it is common belief between the actors in $A$" ($CB_A$), They are defined in the usual way:
\[ EB_A\phi \equiv \wedge_{a\in A} B_a\phi\]
\[ CB_A\phi \equiv \wedge_{i=0}^{\infty} EB^i_A\phi\]
It is exactly the property that several parts of the social practice are common beliefs that makes them work efficiently. 

Given the epistemic operators we can now also define things like the purpose of an action (and a social practice) and expectations within a social practice. The intuitive meaning of the purpose of an action is the reason for which the action is performed. It is part of the intended result of the action and thus indicates that the action is performed with the \emph{goal} of achieving that result. Thus we define the purpose of an action $\alpha$ performed by $a \in A_c$ in a context $c$ as follows:
\begin{gather*} purpose(a,\alpha,c)=\phi \equiv  
CB_{A_c} ((Salient(a,\alpha,c) \\ 
\wedge DO(a,\alpha))\rightarrow Goal(a,\phi) \wedge B_a([\alpha(a)]\phi)
\end{gather*}
If we talk about the general purpose of an action $\alpha$ in context $c$ we mean that whenever any actor performs that action we infer that it was done to achieve that purpose.
\begin{gather*} purpose(\alpha,c)=\phi \equiv \\ 
CB_{A_c} (\forall a\in A: (Salient(a,\alpha,c) \wedge DO(a,\alpha))\rightarrow \\ 
Goal(a,\phi) \wedge B_a([\alpha(a)]\phi)
\end{gather*}
We can make one more abstraction if we let $\alpha$ not be a concrete action but the representation of an abstract action. I.e. $\gamma$=$\alpha\psi$ or in general let $\gamma$=$\cup_{i=1...n}\alpha_i$ then
\[ purpose(\gamma,c)=\phi \equiv \forall \alpha_i: purpose(\alpha_i,c)=\phi\]

\begin{table*}[!th]
\centering
\caption{Social Practice application (A: informal; B: formal)}\label{table:sp}
\begin{tabular}{|M{2.4cm}|M{0.6cm}|M{6.4cm}|M{7.2cm}|N} \hline
\textbf{Social Practice}&&A: Lecture SP (informal)&B: Lecture SP (formal)&\\ \hline
\textbf{Context} && &\\ 
\multicolumn{1}{|r|}{Roles}&$R_{sp}$ &lecturer, student & \textit{l}, \textit{s}\\ 
\multicolumn{1}{|r|}{Actors}&$A_{sp}$&people in the room & $a_1, a_2, ..., a_n$\\ 
 & & $play(a_1,l)$, $\forall a_i, i>1: play(a_i, s)$&\\ 
\multicolumn{1}{|r|}{Resources} &$O_{sp}$&seats, projectors, whiteboard, markers & $o_1,...o_m$\\ 
\multicolumn{1}{|r|}{Affordances} &$\text{Af}_{sp}$ &sit on, present, write with & $\forall o_i$ \emph{afford}$(c_i, \text{\textit{aff}}_j)$, $\text{\textit{aff}}_j \in \{sit, present, write\}$ \\ 
\multicolumn{1}{|r|}{Places} &$\text{Pl}_{sp}$& lecture room & \textit{room}\\ \hline
\textbf{Meaning} && &\\ 
\multicolumn{1}{|r|}{Purpose} &$P_{sp}$&All students learn topic & $purpose(attend,sp)= \newline \forall a\in: plays(a,s), learntopic(a)$\\ 
\multicolumn{1}{|r|}{Promoted Values} &$\text{Pv}_{sp}$& attending lecture \textit{promotes} Self Enhancement, & $\forall i>1: promotes(sp, attend(a_i), SelfEnhance)$\\
&& noisy students \textit{demotes} Respect& $\forall i>1: promotes(sp, noisy(a_i),\neg Respect)$\\ 
\multicolumn{1}{|r|}{Counts-as} &$\text{Co}_{sp}$& student raising hand counts as has a question, & $countsas(sp,raise(s,hand),question(s))$\\ 
&& lecturer raising hand counts as start lecture & $countsas(sp,raise(l,hand),start(lecture))$ \\ \hline
\textbf{Expectations} && &\\ 
\multicolumn{1}{|r|}{Plan Patterns}&$\text{PP}_{sp}$& See Figure \ref{fig:pp}& $\alpha_1(lstarted);$\\
&&& $(\alpha_2(presented)+\alpha_3(q\&a'ed));\alpha_4(lended)$\\ 
\multicolumn{1}{|r|}{Norms} &$N_{sp}$& when lecturer talks students are quiet, & $F(s,talk(l),\neg talk(s))$\\
&& student must raise hand when has a question & $O(s,talk(s),raise(s,hand))$\\
\multicolumn{1}{|r|}{Strategies} &$S_{sp}$& students sit, & $strategy(\top, DO(s, sit))$\\
&& lecturer stops talk when student has question&$strategy(raise(s,hand), DO(lec,stop(talk)))$\\ 
\multicolumn{1}{|r|}{Start Condition} &$\text{Sc}_{sp}$& 9am, students present, lecturer declares start&$strategy(startcondition, DO(all, lecture))$\\ 
\multicolumn{1}{|r|}{Duration} &$D_{sp}$& 11am or lecturer finishes or students gone& $time(11) \vee finish(l) \vee \forall a: play(a,s), leave(a)$\\ \hline
\textbf{Activities} && &\\ 
\multicolumn{1}{|r|}{Possible Actions} &$\text{Ac}_{sp}$& sit, talk, raise hand, ...& $sit(X),talk(X), raise(X,hand),...$\\ 
\multicolumn{1}{|r|}{Requirements} &$\text{Re}_{sp}$& topic of lecture is known& $CB_A(sp,topic(x))$\\ 
&& lecturer: knows topic, can present& $\forall a_i, play(a_i,l)$: \\
&& & $cap(a_i, talk)$, $cap(a_i,raise(a_i,hand))$ \\ 
&& student: has prereq. knowledge for lecture & $\forall a_i, play(a_i,s)$: \\
&& & $cap(a_i, sit)$, $cap(a_i,raise(a_i,hand))$\\ \hline
\end{tabular}
\end{table*}

We have similar definitions for the purpose of complex actions. Let $\gamma \in \sl CA$ then
\begin{gather*}
purpose(a,\gamma,c)=\phi \equiv \\
CB_{A_c} (Salient(a,\gamma,c) \wedge \forall \alpha_i\in act(\gamma):\\ 
DO(a,\alpha_i)\rightarrow Goal(a,\phi) \wedge B_a([\gamma(a)]\phi))
\end{gather*}
Note that this definition only indicates that the whole sequence $\gamma$ leads to $\phi$ and does not restrict exactly which actions are part of that sequence. However, every action is believed to be done because $a$ has the goal to achieve $\phi$. In this way we represent the intuition that each action of the complex action $\gamma$ contributes to the goal $\phi$.\\
We can generalize this definition first by allowing the different actions in $\gamma$ to be done by different actors $a_j$ from a set of actors $A_c$. And we can also abstract altogether from the agency and give the purpose of the complex action $\gamma$. E.g. checking the minutes of the previous meeting at the start of a formal meeting is done for the purpose of creating common ground and agreement about what happened in last meeting.
Let $act(\gamma)=\{\alpha_1;...;\alpha_n\}$ and $A_c=\{a_1,...,a_m\}$ then
\begin{gather*}
purpose(A_c,\gamma,c)=\phi \equiv \\ 
CB_{A_c} (Salient(A_c,\gamma,c) \wedge \forall \alpha_i\exists a_j:\\ 
DO(a_j,\alpha_i)\rightarrow Goal(a_j,\phi) \wedge B_{a_j}([\gamma]\phi)
\end{gather*}
and of course we can talk about the purpose of an abstract sequence of actions $\delta$=$\cup_{i=1...n}\beta_i$:
\[ purpose(\delta,c)=\phi \equiv \forall \beta_i purpose(\beta_i,c)=\phi\]

Using the above definitions we can now give a definition of plan patterns in a social practice $sp$.
Let PP be the smallest set closed under:
\begin{gather*}
\gamma\phi \in PP\\
\gamma_1\phi_1,\gamma_2\phi_2 \in PP \Rightarrow \gamma_1\phi_1 +\gamma_2\phi_2 \in PP \\
\gamma_1\phi_1,\gamma_2\phi_2 \in PP \Rightarrow \gamma_1\phi_1 \&\gamma_2\phi_2 \in PP \\
\gamma_1\phi_1,\gamma_2\phi_2 \in PP \Rightarrow \gamma_1\phi_1 ;\gamma_2\phi_2 \in PP\\
\gamma\phi \in PP \Rightarrow (\gamma\phi)* \in PP
\end{gather*}
Let $\gamma\phi,\gamma_1\phi_1 \in PP$ then we use $\gamma_1\phi_1 \in \gamma\phi$ if $\gamma_1\phi_1$ occurs in $\gamma\phi$.\\
This leads us to the following definition of a plan pattern of a social practice:
\begin{gather*}
planpattern(\gamma\phi,sp) \rightarrow purpose(\gamma\phi,sp)=\phi \wedge \\
\forall\gamma_i\phi_i \in \gamma\phi: purpose(\gamma_i\phi_i,sp)=\phi_i \wedge
\forall\gamma_1\phi_1;\gamma_2\phi_2 \in \gamma\phi:\\ strategy(DONE(A',\gamma_1\phi_1),DO(A''\gamma_2\phi_2),sp)
\end{gather*} 
where $A',A'' \subset A_{sp}$.
This states that plan patterns of a social practice are abstract (sequences of) actions for which the purpose in the context of the social practice is to reach the formula $\phi$ associated with that abstract action. And if the plan pattern specifies two abstract actions in a sequence there is an expectation that when the first part is done (by a subset $B$ of the agents) the second abstract action will be done. (See section 4.3 for more explanation on the strategy relation).
\begin{figure}
\centering
\includegraphics[width=0.9\columnwidth]{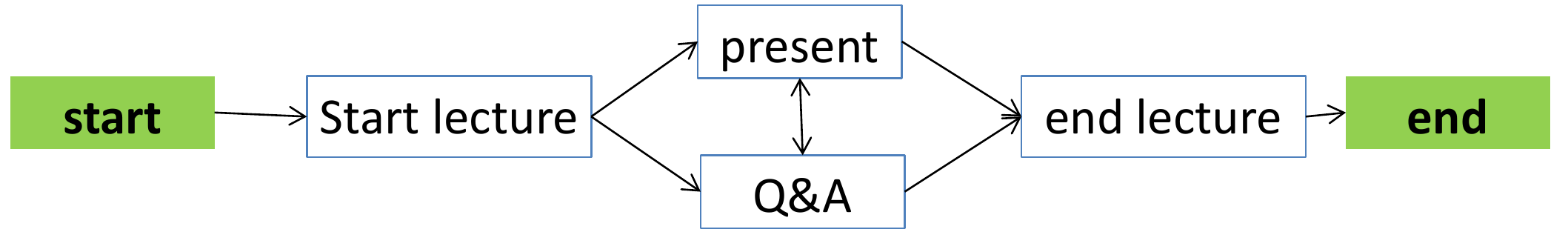}
\caption{Plan Pattern for SP Lecture}
\label{fig:pp}
\end{figure}

\subsection{Meanings}
Besides the above elements that are used in several parts of the formal representation of social practices we also have some more ontological elements that are used to describe meanings and resources. Due to space limitations we will not give complete formal definitions of these elements but refer to other work that gives the (intricate) logical formalization. First of all we define the counts-as relation
\[countsas(sp,\alpha_1,\alpha_2)\]
to mean that performing action $\alpha_1$ in context $sp$ is seen as performing action $\alpha_2$. E.g. the lecturer raising her hand counts as starting the lecture. A further formal definition of this concept of counts-as can be taken from \cite{Grossiphd}.

The second concept that we need is the fact that an action is promoting or demoting a certain value. E.g. students talking during a lecture demotes the value of respect. This concept is used to indicate some abstract social effects of actions. Again this relation is context dependent. We follow \cite{Dignumigpl} and use $play(a,r)$ to indicate that actor $a$ enacts role $r$. Let $sp$ be a social practice $\alpha \in \sl CA$ an (abstract) action, $r$ a role and $v \in V$ a value, 
\[play(a,r) \rightarrow promotes(sp,\alpha(a),v)\]
indicates that if an actor $a$ plays the role $r$ and performs the action $\alpha$ she promotes the value $v$. For a more formal characterization of the promotes relation see \cite{Weidephd}.

Finally we also introduce the concept of affordances here. Affordances are used to describe the type of actions that are normally expected to be performed with an object and can also be used to describe types of objects in an abstract way. E.g. an object that affords to sit (which can be a chair or a bank or ...). The affordance relation is denoted as:
\[\emph{afford}(o,\alpha)\]
where $\alpha \in \sl CA$ and $o$ denotes an object. One could make the affordance relation also context dependent, but as argued in~\cite{oijen2014} a context independent approach suffices.  
%
%

\subsection{Expectations}

Many elements of the social practice can be seen as expressing a kind of expectation. As stated before expectations with social practices come in different forms. The plan patterns can be seen as expectations because they indicate the general patterns of behavior that are expected. However, we also have more specific expectations. Although not all parts of a social practice are fixed there are some points where specific types of actions are expected. E.g. if in a greeting practice one person extends his hand it is expected that the other person shakes the hand.

Let $\gamma \in \sl CA$ and let $B\subset A_{sp}$ be a set of actors (possibly one), then we define strategies (as in ADICO) as follows:
\begin{gather*} 
strategy(\phi, DO(B,\gamma),sp) \equiv\\ 
CB_{A_{sp}}(CB_B\phi) \rightarrow \forall a\in B:CB_{A_{sp}}(DO(a,\gamma(B))
\end{gather*}
Thus if all actors (involved in the social practice $sp$) believe that $B$ believes the condition $\phi$ then they all believe that all actors in $B$ will perform their part of $\gamma$ next. The condition $\phi$ can be of the form $DO(C,\gamma_1)$ or $DONE(C,\gamma_1)$ or a description of other facts. The first two indicate a synchronization and sequencing of actions, while the last represents that it is expected that $B$ takes the initiative to perform some action when a certain state is reached.\\
Note that the above definition entails that if actor $a \in B$ then she believes that if she does her own part of $\gamma$ the rest of $B$ will do their part of $\gamma$. 

As stated before norms are also a kind of expectations. However, instead of a belief that an action will be taken they lead to a normative statement that an action \emph{should} be taken. In the context of social practices, the norms are connected to the roles of the social practice. Thus any actor that enacts a role in the social practice should follow the norms specified for that role. 
Norms are now described as follows:
\[ O(r,\phi,\gamma) \equiv \forall a: play(a,r) \wedge B_a(\phi) \rightarrow O(\gamma(a))\]
\[ F(r,\phi,\gamma) \equiv \forall a: play(a,r) \wedge B_a(\phi) \rightarrow F(\gamma(a))\]
where $O(\gamma)$ and $F(\gamma)$ have the standard dynamic deontic logic semantics.

\section{Representing and Using Social Practices}
In the previous sections we have described all ingredients for representing social practices, which allowed us to fill table \ref{table:sp} completely. Formally a social practice can now be defined as an operator with as arguments all the elements from the table. 
\begin{gather*} socialpractice(sp,R_{sp},A_{sp},O_{sp},Af_{sp},Pl_{sp},P_{sp},\\ 
Pv_{sp},Co_{sp},PP_{sp},N_{sp},S_{sp},Sc_{sp},D_{sp},Ac_{sp},Re_{sp})
\end{gather*}
However in order to achieve the goal of using social practices we should have social practices that can guarantee their fulfillment and give enough expectations to facilitate efficient planning and deliberation.
A social practice $sp$ is useful ($\emph{useful}(sp)$) iff:
\begin{enumerate}
\item for all actions necessary to create at least one trace $t$  that fulfills the plan pattern, norms and all relevant strategies of $sp$ there are objects and actors with capabilities that make those actions possible
%
%
\item if $t$ contains a sequence of actions $\alpha_1(A);\alpha_2(B)$ with $A\not=B$ then $sp$ contains a strategy $strategy(DONE(A,\alpha_1),DO(B,\alpha_2),sp)$ 
\end{enumerate}
The first item indicates that there should be at least one way to fulfill the social practice without violating expectations. The second item indicates that interactions between actors in a social practice are all regulated by strategies and thus do not require explicit synchronization.

In the remainder of this section, we will briefly describe how an actor will use social practices to deliberate about its context and plan its actions. We consider an extremely simplified scenario of attending a lecture on \textit{social practices}. Table \ref{table:sp} describes this social practice along the elements specified in Section \ref{sec:spelements}, providing an informal description and the formal specification using the language introduced in Section \ref{sec:logic}.

We now consider an actor $a$ intending to learn about SP. Assuming this actor to follow a BDI-like architecture, the goal ``know about SP'' would be part of the Intentions of $a$. We further assume that both the social practice described in table \ref{table:sp} and the prerequisite knowledge are part of $a$'s beliefs. Actor $a$ also has the capability to sit leading $a$ to belief that it could fulfill the role of \textit{student}. By using a social practice-based deliberation process, once actor $a$ beliefs to be itself in the context of the \textit{Lecture SP}, it can then form a plan for its participation in the lecture, based on the expectations, meaning and activities the social practice describes for actors of role `student'. This plan is very simple, $a$ sits in the lecture room and leaves at the end of the lecture. While $a$ is sitting it can ask questions when it does not understand the presentation. This is a reactive behavior that is governed by a strategy and a norm. Thus all parts of the plan for the student are given. In fact, based on the plan pattern of sp $a$ can conclude that by fulfilling its part, the desired result will be achieved, given that it expects all other agents to fulfill their own parts.
For the lecturer there is a bit more planning left. The lecturer has to plan the sequence of actions that will instantiate the abstract "present" action.\\
Note also that the current social practice does not require the students to actually pay attention. In fact, the (expected) social effects of actions is an important issue but of the scope of this paper. Social effects are not visible but can be assumed given the social practice. E.g. if no questions being asked can be taken to mean that students understand the subject so far, and therefore no explicit checking is needed.

\section{Conclusions}
In this paper we gave a first formalization of social practices that aims at support actors planning  social interactions. We have briefly shown that {\emph useful} social practices can simplify the amount of actions that have to be planned by an actor and that it can make assumptions about interactions with other actors in the social practice without explicitly having to coordinate.

Due to space limitations we cannot present the full logic nor discuss the formal properties of social practices that follow from its formalization. We leave these for future work. Another important area for future work is the development of learning mechanisms that enable efficient matching of sensing information and beliefs to the components of a social practice description.

\clearpage

\newpage
\bibliographystyle{named}
\bibliography{social}

\end{document}